\documentclass{article}

\usepackage{PRIMEarxiv}

\usepackage[utf8]{inputenc} % allow utf-8 input
\usepackage[T1]{fontenc}    % use 8-bit T1 fonts
\usepackage{hyperref}       % hyperlinks
\usepackage{url}            % simple URL typesetting
\usepackage{natbib}
\usepackage{amsmath}
\usepackage{dsfont}
\usepackage{cleveref}
\usepackage{booktabs}  
\usepackage{amsfonts}       % blackboard math symbols
\usepackage{nicefrac}       % compact symbols for 1/2, etc.
\usepackage{microtype}      % microtypography
\usepackage{lipsum}
\usepackage{fancyhdr}       % header
\usepackage{graphicx}       % graphics
\graphicspath{{media/}}     % organize your images and other figures under media/ folder
\usepackage{graphicx}
\usepackage{subcaption}
\usepackage{longtable}
\usepackage{authblk}
\usepackage{booktabs}    % Improves table appearance
\usepackage{multirow}    % Allows multi-row cells
\usepackage{color}
\usepackage{hyperref}

\hypersetup{
     colorlinks   = true,
     citecolor    = black,
     linkcolor    = blue,
}

\title{Human Preferences in Large Language Model Latent Space: A Technical Analysis on the Reliability of Synthetic Data in Voting Outcome Prediction
}
\author[1,5]{Sarah Ball$^*$}
\author[2,4]{Simeon Allmendinger$^*$}
\author[1,3,5]{Frauke Kreuter}
\author[2,4]{Niklas Kühl}

\affil[1]{Ludwig-Maximilian-University Munich}
\affil[2]{University of Bayreuth}
\affil[3]{University of Maryland}
\affil[4]{Fraunhofer Institute for Applied Information Technology FIT}
\affil[5]{Munich Center for Machine Learning MCML}

\date{} % Removes date

\begin{document}
\maketitle

%\footnotetext{Sarah Ball and Simeon Allmendinger contributed equally to this work.}

% Temporarily change footnote symbol to an asterisk
\renewcommand{\thefootnote}{\fnsymbol{footnote}}
\footnotetext[1]{Equal contribution.}
\renewcommand{\thefootnote}{\arabic{footnote}} % Reset to default numbering

\begin{abstract}
Generative AI (GenAI) is increasingly used in survey contexts to simulate human preferences. While many research endeavors evaluate the quality of synthetic GenAI data by comparing model-generated responses to gold-standard survey results, fundamental questions about the validity and reliability of using LLMs as substitutes for human respondents remain. Our study provides a technical analysis of how demographic attributes and prompt variations influence latent opinion mappings in large language models (LLMs) and evaluates their suitability for survey-based predictions.
Using 14 different models, we find that LLM-generated data fails to replicate the variance observed in real-world human responses, particularly across demographic subgroups. In the political space, persona-to-party mappings exhibit limited differentiation, resulting in synthetic data that lacks the nuanced distribution of opinions found in survey data. Moreover, we show that prompt sensitivity can significantly alter outputs for some models, further undermining the stability and predictiveness of LLM-based simulations.
As a key contribution, we adapt a probe-based methodology that reveals how LLMs encode political affiliations in their latent space, exposing the systematic distortions introduced by these models. %By linking LLM mechanisms with practical considerations, we demonstrate that current LLMs are unreliable predictors for opinion research or decision-making simulations. 
Our findings highlight critical limitations in AI-generated survey data, urging caution in its use for public opinion research, social science experimentation, and computational behavioral modeling.

\end{abstract}

%%%%%%%%%%%%%%%%%%%%%%%%%%%%%%%%%%%%%%%%%%%%%%%%%%%%
\section{Introduction}

With the release of ChatGPT in November 2022, the world has seen a spike in interest in large language models (LLMs). Many academic disciplines, as well as the business world, wonder if and how they can integrate LLMs to their benefit. One emerging---and highly debated---topic is the usage of LLMs for (public) opinion research. The idea is that one can leverage LLMs to substitute for surveying humans. Yet, the question remains as to how valid and reliable it is to substitute humans with LLMs. Previous research mainly focuses on comparing LLM predictions based on personas to a gold standard survey prediction for these personas. The results of such analyses are mixed~\citep{argyle2023out, %durmus2023towards, 
kim2023ai}, revealing various problems, e.g., prediction instability that occurs with slight formulation changes in the prompt~\citep{bisbee2023synthetic} and performance differences across national and linguistic contexts~\citep{von2024united}. While such approaches might give first insights into how well LLMs can predict general questions of interest, we lack a deeper understanding of how ``opinion formation'' works on a \textit{technical} level in LLMs and how reliable the resulting synthetic data is for answering human-related questions of interest. Based on this, our article addresses two central research questions: 

\begin{enumerate}
    \item[\textbf{\textit{RQ1:}}] How well does LLM-generated synthetic data mimic the distribution of human answers in survey-like questions for different demographic subgroups in their latent space?
    \item[\textbf{\textit{RQ2:}}] How is prompt instability reflected in the models' latent space? 
\end{enumerate}
 
To address these questions, we focus on the use case of predicting election outcomes with LLMs in the German multi-party context. The election context is chosen not only for its societal relevance and its popularity as a testbed in recent research on LLM-human substitutability~\citep{argyle2023out,von2024united,yu2024large}, but also because elections are commonly used for evaluating the quality of survey data across different vendors or data collection modes, providing a rare benchmarking opportunity in survey research. We further choose the German multi-party context as it allows for multiple party comparisons, increasing the robustness of our results. 

In our experiments, we analyze the latent space of LLMs, focusing on mechanistically understanding persona-to-party mappings.
To do so, we develop a probe-based methodology to systematically identify model-specific value vectors---Multi-Layer Perceptrons (MLPs)---associated with political affiliations.
This allows us to examine how demographic attributes---such as age, gender, and ideological leaning---interact with latent political structures within LLMs.
Our results reveal that \textbf{LLMs fail to replicate the entropy observed in real-world survey data}, as their persona-to-party mappings exhibit low differentiation across demographic subgroups.
We further explore prompt sensitivity by first replicating previous findings that small meaning-preserving variations in persona descriptions can alter voting predictions, underscoring the instability of LLM-generated survey data. Next, we demonstrate that, \textbf{for certain models, higher entropy in the persona-to-party mapping correlates with increased prompt sensitivity}. However, we also observe the opposite relationship in other models.

Overall, our study provides a technical foundation to assess the usability and reliability of synthetic LLM data, exposing fundamental limitations that practitioners must address before relying on LLMs for public opinion research, social science experimentation, and computational behavioral modeling. We preregistered our study on the Open Science Framework\footnote{Due to time constraints, we reduced the number of parameters to consider in our study for the preprint at hand (\hyperlink{https://osf.io/2hmb5}{OSF}).}, and the code is available at GitHub\footnote{Codebase in \hyperlink{https://github.com/SimeonAllmendinger/opinion_representation_in_llms.git}{GitHub Repository}.}.

%i) our probes can identify party value vectors of the LLM with high certainty, ii) persona attributes, such as gender or age, correlate differently with political subscriptions embedded in the value vectors, and iii) varying textual phrases describing persona attributes heavily impact the results. 

%%%%%%%%%%%%%%%%%%%%%%%%%%%%%%%%%%%%%%%%%%%%%%%%%%%%
\section{Related Literature}
\textbf{Using LLMs as substitutes for humans.} The advent of large language models (LLMs) has sparked significant interest regarding their potential to serve as substitutes for human respondents~\citep{argyle2023out}. This question is especially relevant for survey researchers in the social sciences, who are investigating whether responses generated by LLMs can reliably resemble those provided by humans in surveys~\citep{argyle2023out, bisbee2023synthetic, dominguez2025questioning, park2024generative, qu2024performance, vonderheyde2024uniteddiversitycontextualbiases, wang2024large}. Similar inquiries have emerged in fields such as market research~\citep{brand2023using, sarstedt2024using}, annotation tasks~\citep{tornberg2023chatgpt, ziems2024can}, experiments in psychology and economics~\citep{aher2023using, xie2024can}, and human-computer-interaction~\citep{hamalainen2023evaluating, tornberg2023chatgpt}, among others. The findings from these investigations are mixed. Some studies suggest that LLMs can reasonably approximate the average outcomes of human surveys~\citep{argyle2023out, bisbee2023synthetic, hamalainen2023evaluating, tornberg2023chatgpt, brand2023using, xie2024can}, while others highlight significant limitations, particularly in their inability to accurately represent the opinions of diverse demographic groups~\citep{santurkar2023whose, vonderheyde2024uniteddiversitycontextualbiases, sarstedt2024using, qu2024performance, dominguez2025questioning}. However, a common limitation across these studies is their focus on surface-level comparisons, i.e., matching LLM output to human survey responses without delving into the underlying mechanisms of how opinions are encoded and represented within the models' latent spaces. We address this gap by studying how personas are mapped to opinions as well as what the inherent limitations are in eliciting specific knowledge from these models. %Addressing these questions at a deeper, technical level is a central aim of our research.

\textbf{Prompt sensitivity.} By systematically introducing subtle changes to the prompt format, previous studies have shown that LLM output is highly sensitive to prompt changes, thereby influencing downstream evaluations~\citep{leidinger2023language, mizrahi2023state, chatterjee2024posix, voronov2024mind, zhuo2024prosa}. Articles most closely related to our study are~\citet{sclar2023quantifying} and~\citet{zhu2023promptrobust}, which focus on explaining LLM prompt sensitivity next to establishing that it exists. \citet{sclar2023quantifying} analyze how changes in the formatting of the prompt without semantic changes lead to large performance differences. They further show that prompt embeddings of different but equivalent formats are distinguishable using a trained classifier, implying that prompt formats transform the output probability distribution, yielding different predictions. \citet{zhu2023promptrobust} design ``attacks" on the character, word, sentence, and semantic level to mimic user errors. They again find significant performance differences induced by the subtle prompt changes. The study further examines why LLMs are vulnerable to adversarial inputs by analyzing their attention weights when processing both clean and adversarial prompts. The findings indicate that these adversarial prompts redirect the model's attention towards the altered elements, leading to incorrect responses.
We build on these interpretability approaches and offer a different perspective of how persona-to-party mapping entropy is related to prompt sensitivity.  

%%%%%%%%%%%%%%%%%%%%%%%%%%%%%%%%%%%%%%%%%%%%%%%%%%%%
\section{Models and Data}
\label{sec:modelsanddata}

\textbf{Model selection.} For our experiments we use both base and aligned models of different model families and sizes, see ~\Cref{tab:llm_overview}. These models are developed by teams across the world and fulfill the white-box criteria, which is a requirement for studying their latent space. 

\begin{table}[h]
    \centering
    \caption{Overview of LLM models used in the experiments.}
    \begin{tabular}{lll l}
        \toprule
        \textbf{Family} & \textbf{Size} & \textbf{Model} & \textbf{Reference} \\
        \midrule
        Llama 3.2 & 3B  & Llama-3.2-3B-Instruct  & ~\citet{meta2024llama32}  \\
                  & 3B  & Llama-3.2-3B  & ~\citet{meta2024llama32} \\
        Llama 3.1 & 8B  & Llama-3.1-8B-Instruct & ~\citet{meta2024llama31} \\
                  & 8B  & Llama-3.1-8B  & ~\citet{meta2024llama31}  \\
        Llama 3   & 8B  & Llama-3-8B-Instruct  & ~\citet{meta2024llama3}  \\
                  & 8B  & Llama-3-8B  & ~\citet{meta2024llama3}  \\
        Llama 2   & 7B  & Llama-2-7b-hf  & ~\citet{touvron2023llama2}  \\
                  & 7B  & Llama-2-7b-chat-hf  & ~\citet{touvron2023llama2} \\
        Mistral   & 7B  & Mistral-7B-v0.1  & ~\citet{jiang2023mistral7b}  \\
                  & 7B  & Mistral-7B-Instruct-v0.1  &~\citet{jiang2023mistral7b}  \\
        Gemma     & 7B  & Gemma-7b-it  & ~\citet{gemma}  \\
                  & 7B  & Gemma-7b  & ~\citet{gemma} \\
        Qwen      & 7B  & Qwen2.5-7B  & ~\citet{qwen2025qwen25technicalreport} \\
                  & 7B  & Qwen2.5-7B-Instruct  & ~\citet{qwen2025qwen25technicalreport} \\
        \bottomrule
    \end{tabular}
    \label{tab:llm_overview}
\end{table}

\textbf{Real world comparison.} In order to compare our model predictions to real data, we use the German Longitudinal Election Study (GLES)~\citep{GESIS_GLES}. This representative survey captures insights about German citizens' political attitudes, preferences and behaviours and is a widely used gold standard~\citep{Schmitt-Beck2010}. As a baseline, we choose the  cross-sectional survey of the year 2021\footnote{In future iterations of this manuscript, we will repeat the comparison with 2025 data.}, for which the GLES asks about voting decisions in the respective federal elections and also captures our variables of interest. To obtain a representative comparison baseline for our LLMs we weight the data with a socio-demographic weight that aligns the distributions to the marginal distributions of the 2021 Microcensus. 

\textbf{Personas.} For the construction of the personas, we follow previous literature~\citep{von2024united} by combining political science theory for identifying voting predictors and representative surveys for extracting plausible values for these predictors. Hence, our personas are both theory- and data-driven. Concretely, we select the variables \texttt{age, gender, education, hhincome, employment, political orientation}, and whether a person \texttt{lives} in East or West Germany and combine them in a prompt\footnote{See Appendix–\Cref{tab:variables-and-groups} on details for the values of the specific variables.}. We vary the values for the different variables while holding the prompt structure fix. Furthermore, to account for LLM models' prompt sensitivity, we paraphrase the prompts. Thus, an example persona instantiated via an LLM prompt reads as follows: 

\begin{quote}
    \textit{I am \{age\} years old and \{gender\}. I have \{education\}, a \{hhincome\} household net income per month, and I am \{employment\}. Ideologically, I lean towards the position \{left leaning\}. I live in \{east germany\}.
    If the elections were held in \{year of election\}, which party would I vote for? I vote for the party ...}
\end{quote}

\textbf{Probe generation.} To train our probe, we use the German ``Wahl-o-Mat'' data~\citep{wahlomat_bpb}. The Wahl-o-Mat is an online questionnaire, which consists of short political statements based on party manifestos to which interested citizens can give their agreement (strong agree to strong disagree). Based on the user's answers, the tool provides a voting recommendation. For all short political statements that users see, each party provides an opinion to give more context to the question of interest. We extract this opinion for each Wahl-o-Mat item for German and European elections from 01/2021 until 12/2024. The parties of interest are, in alphabetical order, the Alternative für Deutschland (AfD), Christlich Demokratische Union (CDU), Freie Demokratische Partei (FDP), Sozialdemokratische Partei Deutschland (SPD), Bündnis 90/Die Grünen (GRÜNE), and DIE LINKE. 

%%%%%%%%%%%%%%%%%%%%%%%%%%%%%%%%%%%%%%%%%%%%%%%%%%%%
\section{Methodology}

\begin{figure}
    \centering
    \includegraphics[width=0.9\linewidth]{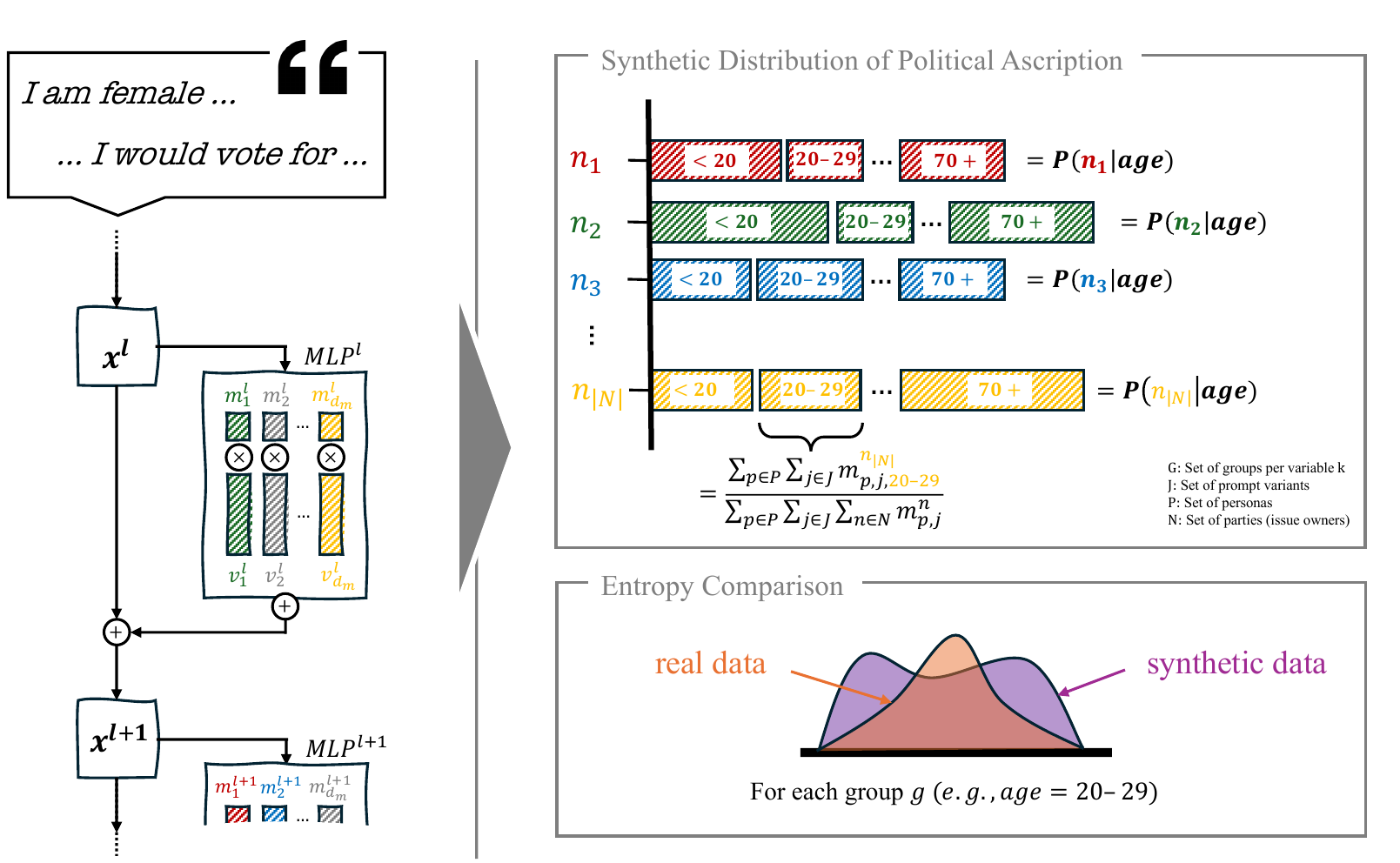}
    \caption{Method overview for comparing latent space persona-to-party mappings with real world voting distributions.}
    \label{fig:method}
\end{figure}

Understanding how LLMs encode and generate synthetic survey responses necessitates to investigate persona-to-party mappings within the models' latent space. Building upon prior research, our methodology integrates trained probes to systematically identify model-specific representations of political ascriptions, thereby offering insights into the underlying value vectors.
As depicted in~\Cref{fig:method}, our methodology depicts how LLM architectures encode voting preferences compared to historical human preferences from GLES.

\subsection{Technical Preliminaries}

Each transformer model~\citep{vaswani2017attention} consists of transformer blocks in which multihead-attention (MHA) and multilayer perceptrons (MLP) update the residual stream representation ($x_i^l$) in each layer $l$ to obtain an updated representation $x_i^{l+1}$ (bias terms omitted for brevity)~\citep{elhage2021mathematical}:

\begin{equation}
    x_i^{l+1} = x_i^l + MLP^l(x_i^l + MHA^l(x_i^l)), l=1,2,...,L
\end{equation}

%The residual stream at the last layer then gets projected to the vocabulary space to obtain a distribution over the vocabulary (after softmax): 

Based on~\citep{geva2022transformer} we can further decompose each MLP into two linear transformations (note that we write $x_i^l = x^l$ for brevity): 
\begin{equation}
MLP^l(x^l)=f(W_K^lx^l)W_V^l,
\end{equation}
where $f$ is a non-linear activation function and $W_K^l ,W_V^l \in \mathbb{R}^{d_{mlp}\times d}$. Hence, each value vector $v_i^l$ in column $i$ of $W_V^l$ is weighted by a vector of coefficients $m^l := f(W_K^lx_i^l) \in \mathbb{R}^{d_m}$. Noting $k_i^l$ as the key vector of row $i$ in  $W_K^l$, one can write:
\begin{equation}
    MLP^l(x_i^l)=\sum^{d_m}_{i=1}f(x^l_i ·k^l_i)v_i^l =\sum^{d_m}_{i=1}m^l_iv_i^l
\end{equation}
From this equation,~\citep{geva2022transformer} interpret an MLP update to the residual stream as sub-updates, consisting of weighted value-vectors. They further show that in each sub-update, $v_i^l$ either de- or increases the probability of a token $t$ to be generated: 

\begin{equation}
    p(t|x^l +m^l_iv_i^l,E) \propto exp(e_t \cdot x^l)\cdot exp(e_t \cdot m^l_iv_i^l), 
    \label{eq:token_probability}
\end{equation}

where $e_t$ is the embedding of token $t$ and $E$ the embedding matrix used to generate the first embedding of this token. Importantly, when $e_t \cdot m^l_iv_i^l < 0$, the probability of $t$ decreases and vise versa if $e_t \cdot m^l_iv_i^l > 0$. Furthermore, $e_t \cdot v_i^l$ is static and does not depend on the input, which is why the impact of the input is determined by the scaling $m^l_i$, which itself is determined by the key vector, $k_i^l$ and the residual stream representation $x^l$. Given this decomposition for our case at hand, we are first interested in identifying these ``static'' value vectors, which most increase the likelihood of outputting a token related to a party. We then analyze the scaling $m_i^l$ of these identified value vectors, induced by the personas, which are our inputs of interest.

\subsection{Constructing a probe for identifying party-related MLP value vectors}

We aim to extract value vectors from the intermediate layers of LLMs, as these layers capture conceptual structures and high-level semantic representations more effectively than final layers, which are predominantly specialized for next-token prediction~\citep{panickssery2023steering}.
By focusing on these layers, we seek to uncover how specific residual stream patterns correlate with political biases and party affiliation in LLMs.
To achieve this, we train linear probes that predict the party on the basis of the residual stream $\bar{x}^l$ of layer $l$.
Similar to~\citep{lee2024mechanistic}, these probes help identify value vectors that promote tokens linked to specific partys. 
The probes are trained as binary classifiers, distinguishing between residual streams corresponding to a specific party $n$ $(y=1)$ and all others $(y=0)$. The training loss is weighted to mitigate class imbalance:

\begin{equation}
    \mathcal{L} = -w_1 y \log(\hat{y}) - (1 - y) \log(1 - \hat{y})
\end{equation}

where $w_1$ represents the weight for the positive class. The probe function follows:

\begin{equation}
    P(n|\bar{x}^l) = \text{softmax}(W_{n}\bar{x}^{l}), \quad W_{n} \in \mathbb{R}^d_m, \quad l \in [0.6L,0.9L],
\end{equation}

where $W_{n}$ represents the learned parameters, $n$ symbolizes a party from $N = \{n_1, n_2,\dots\}$, and $\bar{x}^{l}$ is the mean residual stream of a selected layer $l$ from the interval $[0.6L,0.9L]$.
The model consists of a linear layer with dropout, optimized using Adam with cosine annealing.
The training data consists of opinion statements.
Each statement is paired with the corresponding party opinion to construct prompts.
The language model aims to predict the party from the statement-opinion pairs, and residual streams are recorded at all layers and token positions ($=$ sequence).
After training, we extract value vectors per party by identifying MLP weights most aligned with the trained probe weights.
These vectors are selected based on cosine similarity, where $W_{\text{probe}}$ represents the trained probe weights and $v_{i,l}$ denotes the value vector:
\begin{equation}
    \cos(\theta_{i}^{l}) = \frac{W_{\text{probe}} \cdot v_{i}^{l}}{\|W_{\text{probe}}\| \|v_{i}^{l}\|}, \quad i \in [1, d_m], \quad l \in [1, L].
\end{equation}

We define the set of top 20 value vectors per party $n$ as:
\begin{equation}
    \hat{V}^n = \{ v_i^l \mid \cos(\theta_{i}^{l}) \text{ is among the top 20 for } i \in [1, d_m], \text{ and } l \in [1, L] \}.
\end{equation}
The selection criterion ensures that only the 20 most aligned value vectors are retained per layer, as we observe a drop in cosine similarity beyond this threshold.

\subsection{Analyzing the mapping between personas and the identified party-related value vectors}

Personas are defined by key attributes (such as age, gender, and political attitudes) with systematically varied values and paraphrased prompt variants to ensure robustness.
These controlled inputs allow us to analyze how different demographic and ideological configurations affect model residual streams in response to political prompts.
To investigate the interaction between personas and the identified value vectors $v_i^l$, we compute their residual stream of responses to persona prompts.
Specifically, we measure the contribution of scaling factors $m_p \in \mathbb{R}^N$ (cf. \cref{eq:token_probability}) by evaluating the residual stream across all layers for each value vector $v_i^l$.
The scaling factor of each persona prompt is computed as:
\begin{equation}
    m_{p} = \sum_{i,l} m_i^l \cdot \frac{\cos(\theta_i^l)}{\sum_{i,l} \cos(\theta_i^l)}
    \cdot \mathds{1}\{ v_i^l \in \hat{V} \}
    , \quad i \in [1, d_m], \quad l \in [1, L],
\end{equation}
where $m_i^l$ represents the scaling contribution of value vector $v_i^l$ located at layer $l$ and model dimension $i$, and $\cos(\theta_i^l)$ denotes its alignment with the learned probe weights.
Using this approach, we derive scaling behavior $m_p^n$ for each persona $p$ and party $n$ and have an angle to quantify how different demographic and ideological configurations influence residual streams within LLMs.

\subsection{Comparing survey and LLM distributions}
\label{subsec:psi-distribution}

To compare the characteristics of the LLM persona-to-party mapping with historical human preferences, we calculate the normalized entropy $H_{norm}(\psi)$ for the distribution $\psi$ aggregated over all personas $p \in P = \{p_1, p_2,\dots\}$ and prompt variants $j \in J = \{j_1, j_2,\dots\}$.

\begin{equation}
    H_{norm}(\psi) = \frac{H(\psi)}{H_{max}(\psi)} = \frac{- \sum_{n \in N} p(n) \log_2 p(n)}{\log_2 N},
\end{equation}

with $\psi$ defined as:

\begin{equation}
    \psi = \frac{\sum_{p\in P}\sum_{j \in J} m_{p,j} }{\sum_{p\in P}\sum_{j \in J}\sum_{n \in N} m_{p,j}^n} = \frac{m}{\sum_{n \in N} m^n}, \quad \psi \in [0,1]^N.
\end{equation}

While traditional LLM-based political inference often focuses only on next-token prediction, this value-based distribution extends beyond single-token outputs, capturing a full probability distribution over all parties $n$ for each persona $p$. This allows for a more structured comparison with real-world survey data, as it reflects not just the most likely choice, but the entire spectrum of voting preferences inferred from the LLM's internal representations. By comparing $\psi$ with observed human voting distributions, we can assess whether LLMs replicate the variance and entropy observed in real-world political behavior or exhibit systematic biases in persona-to-party mappings.

\subsection{Analyzing prompt sensitivity}
The normalized entropy $H_{\text{norm}}(.)$ can be decomposed by considering the distribution $\psi_g$ for a specific group $g \in G = \{\text{female, college,} \dots\}$, which defines a subset of personas $P_g \subseteq P$. Thus, $H_{\text{norm}}(\psi_g)$ characterizes the entropy behavior within the persona distribution for a given group. Similarly, the distribution $\psi_j$ is constructed based on a subset of personas $P_j \subseteq P$ corresponding to prompt variant $j$, where each subset satisfies $P_{j,g} = P_g \cap P_j \neq \emptyset$, $\forall g, j.$

To assess how entropy varies across different prompts $j$, we examine the relationship between entropy and prompt sensitivity using the Wasserstein distance $W(\psi_{j,g},\bar{\psi}_{j,g})$, which measures the discrepancy between the distribution $\psi_{j,g}$ for prompt variant $j$ and its barycenter $\bar{\psi}_{j,g}$:

\begin{equation}
    \bar{\psi}_{j,g} = \frac{1}{\bigm|J\bigm|} \sum_{j \in J} \psi_{j,g}, \quad g \in G.
\end{equation}

The Wasserstein distance quantifies the minimal effort required to transform one distribution into another in terms of probability mass transport. This allows us to interpret $W(\psi_{j,g},\bar{\psi}_{j,g})$ as a proxy for prompt sensitivity. It captures the extent to which persona distributions shift across different prompt formulations. A higher Wasserstein distance indicates greater instability, meaning that minor variations in prompts lead to significantly different latent representations. To formally assess this effect, we regress $W(\psi_{j,g},\bar{\psi}_{j,g})$ on the normalized entropy $H_{\text{norm}}(\psi_{j,g})$, evaluating how prompt-induced variation correlates with entropy within persona distributions.

%%%%%%%%%%%%%%%%%%%%%%%%%%%%%%%%%%%%%%%%%%%%%%%%%%%%
\section{Results}

We analyze how LLMs model persona-to-party mappings and compare their voting distributions to real-world election data. First, we examine the entropy of persona voting distributions to assess whether models capture variability in political preferences. Next, we compare the predicted voting distributions to observed election outcomes, highlighting systematic biases. Finally, we investigate prompt sensitivity by measuring how variations in phrasing affect model predictions using the Wasserstein Distance as a proxy for prompt sensitivity.

\subsection{Comparing Persona-to-party mapping and real world distribution}
\begin{figure}[h]
    \centering
    \includegraphics[width=\linewidth]{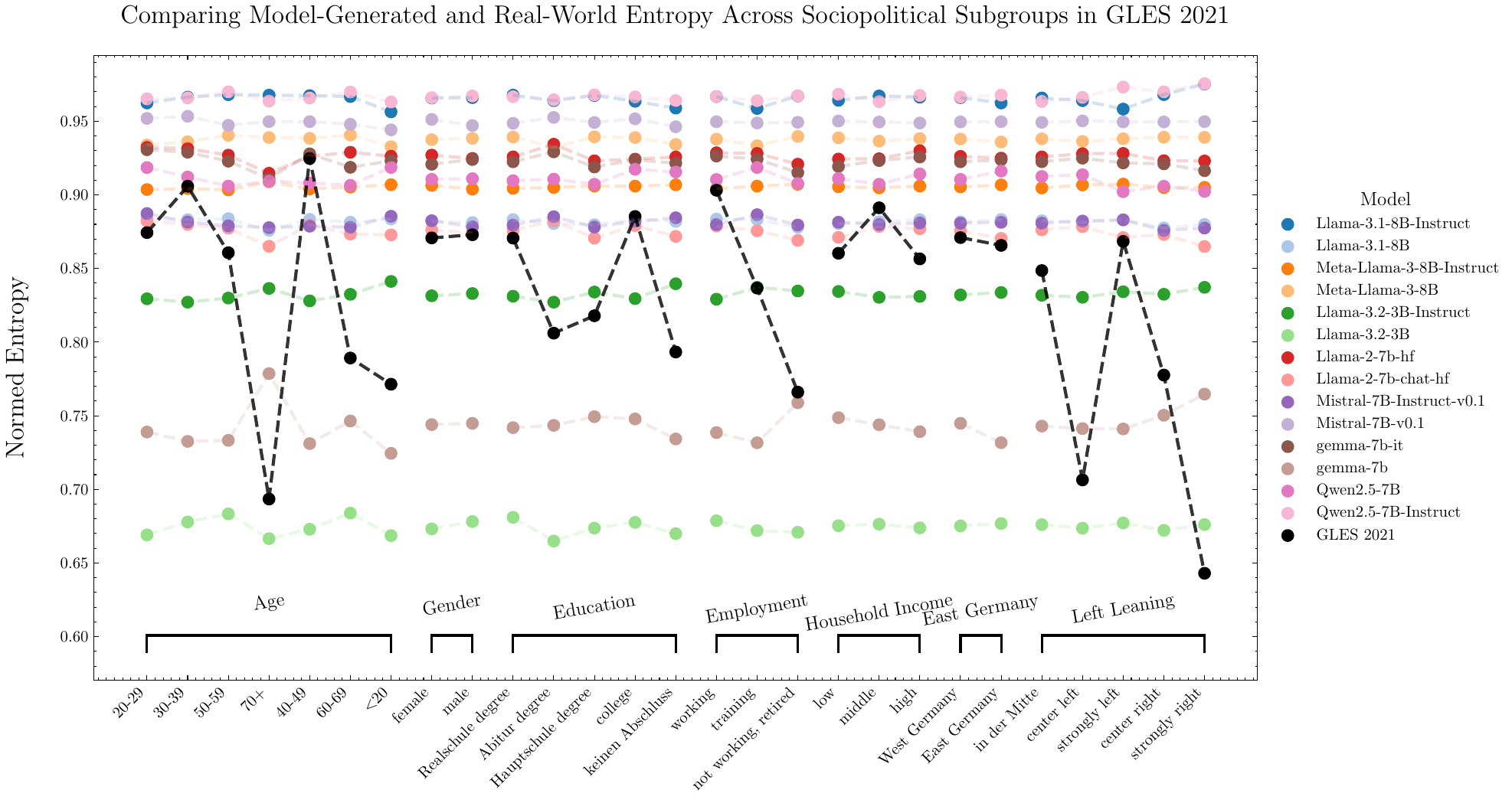}
    \caption{Comparison of the entropy of voting outcomes across different sociopolitical subgroups (e.g., female) as predicted by LLMs versus the real-world entropy observed in the GLES data. Higher entropy indicates greater uncertainty or diversity in political preferences within a subgroup. %The analysis highlights how models lack to capture the variability in voter behavior compared to empirical election data.
    }
    \label{fig:entropy_all_models}
\end{figure}

In our first set of experiments we compare the persona-to-party mappings in the LLMs' latent space to the real world voting distributions by looking at the distributions' normalized entropy. \Cref{fig:entropy_all_models} shows that overall, the entropy values are at a similar level across the different variable groups for the LLMs but not so much for the GLES data. For instance, entropy values range between $0.96$ and $0.98$ no matter which persona we give to \textit{Qwen2.5-7B-Instruct}. This pattern of little variation across personas applies to most of the models with higher entropy values than GLES. For the GLES data, the entropy values differ within but also between groups. For instance, overall entropy for \texttt{education} is higher than for \texttt{left leaning}. For the latter, we also have entropy values ranging between $0.64$ and $0.88$, representing a wider range. Interestingly, the age group of \texttt{70+} induces a more noticable change in entropy to the otherwise stable values for \textit{Gemma-7B}. However, while entropy decreases in the GLES data compared to other age groups, it increases for the model. 

In general we observe, that there is a difference in entropy levels between the models ranging from entropy values as low as $0.66$ to as high as $0.98$. 
While most models' entropy values are above those of the real world baseline, \textit{Llama-3.2-3B} and \textit{Gemma-7b} tend to have lower entropy values. This implies that these models have persona-to-party mappings that are more distinct than in the GLES. For the variables \texttt{gender}, \texttt{hhincome}, and \texttt{East Germany}, \textit{Mistral-7B-Instruct-v0.1} and \textit{Llama-3.1-8B} closely match the GLES baseline. However, matching entropy values do not imply matching voting predictions.

\begin{figure}[h]
    \centering
    \includegraphics[width=\linewidth]{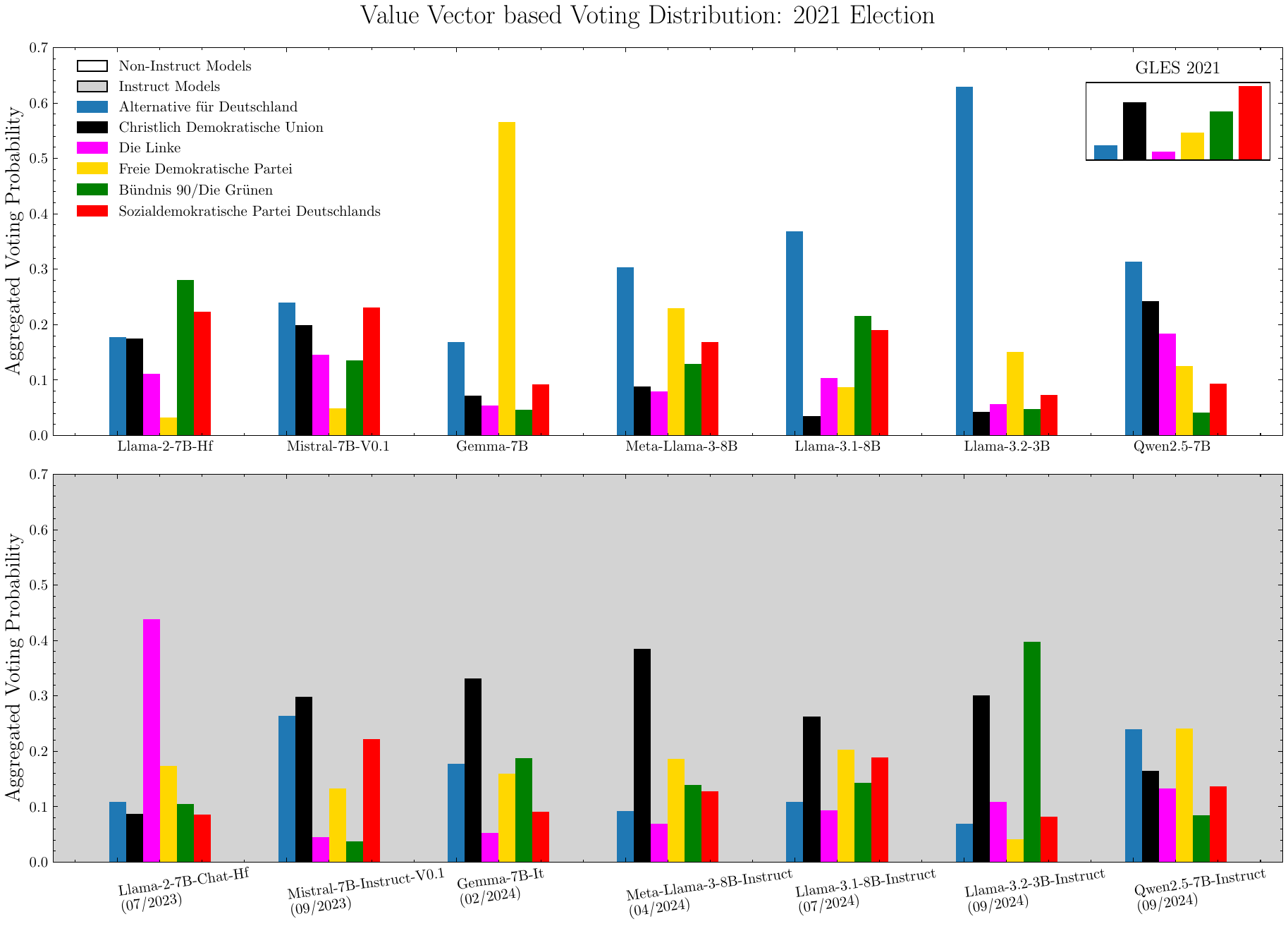}
    \caption{Value vector based distribution $\psi$ in the latent space for election year 2021 aggregated over the different personas according to their occurrence in the representative survey GLES.
    The top panel depicts base models, which show a tendency towards the right-populist AfD. The lower panel shows aligned models, where voting distributions shift towards CDU and other democratic, left-leaning parties. \textit{Qwen2.5-7B} is closest to real-world outcomes.}
    \label{fig:voting_distribution}
\end{figure}

In ~\Cref{fig:voting_distribution} we compare the voting outcome distribution for the different parties as predicted by the value vector based distribution $\psi$ in the latent space (see~\Cref{subsec:psi-distribution}). The distribution is weighted by the occurrence of the personas and their weighting in the GLES. The top panel depicts base models, while the lower panel shows the voting results for the aligned models. For the base models, we observe a clear trend towards the right, populist party AfD, except for \textit{Llama-2-7B-Hf}, \textit{Mistral-7B-V0.1} and \textit{Gemma-7B}. These models also predict more center-left parties like the SPD and GRÜNE or liberal parties like the FDP. In contrast, the aligned models' voting distributions $\psi$ mainly shift in favor of CDU, but also all the other democratic more left-leaning parties. The model closest to the real world outcome distribution is \textit{Qwen2.5-7B} having the smallest Wasserstein Distance of $0.0127$.

We repeat our analyses by asking the model to select a party for a specific persona given the election was \textit{tomorrow}. The results in~\Cref{ap:entropies} indicate similar entropy and voting distribution patterns as with the election year 2021. In addition, we provide further details on how different persona groups trigger different value vectors in~\Cref{ap:regressionmapping}, which provides the basis for our entropy analysis.

\subsection{Prompt Sensitivity}

\begin{figure}[h]
     \begin{subfigure}[b]{0.45\textwidth}
         \centering
        \includegraphics[width=\textwidth]{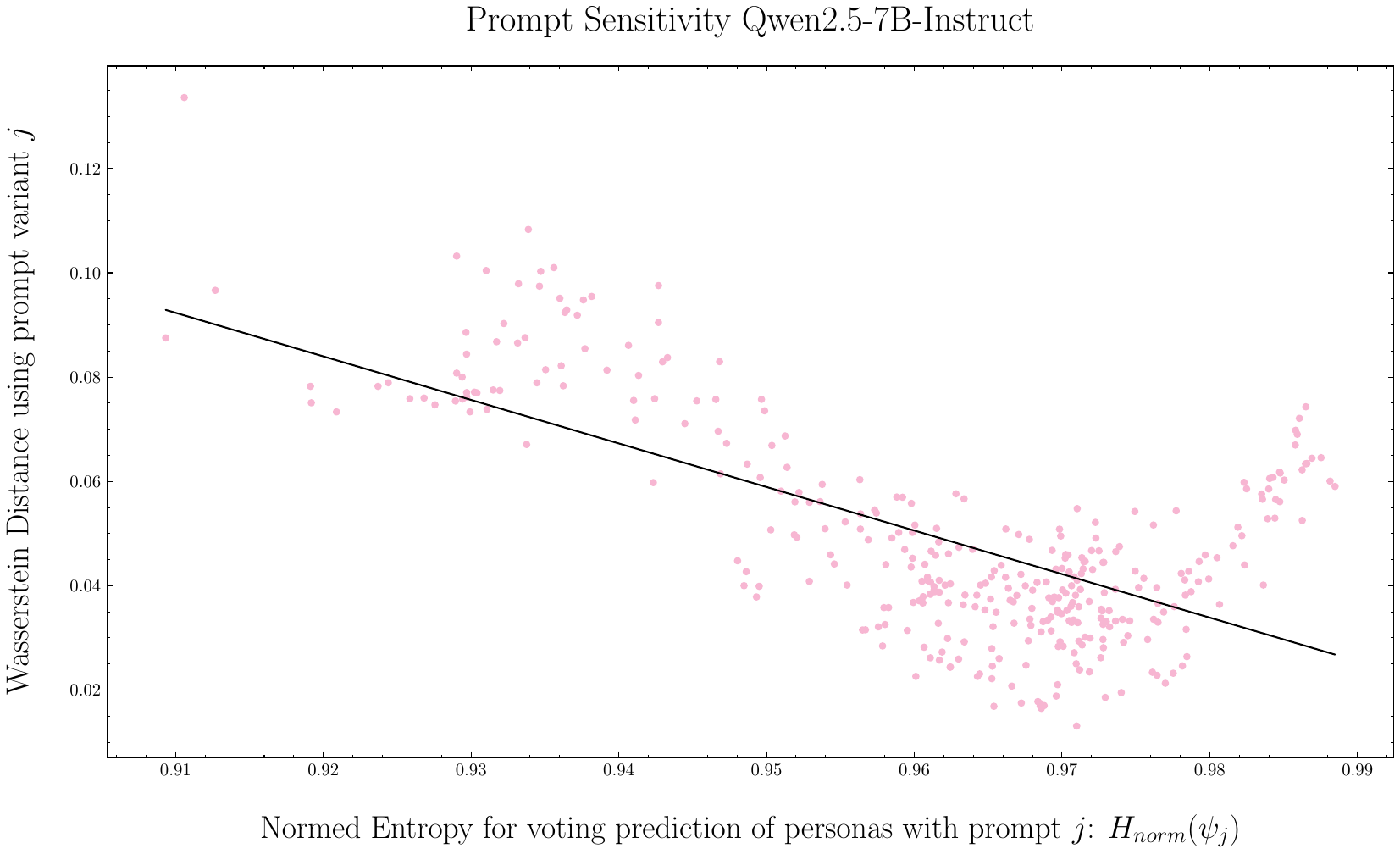}
        \caption{Negative relationship for Qwen2.5-7B-Instruct: higher entropy reduces prompt sensitivity.}
        \label{fig:prompt-sensitivity-qwen}
    \end{subfigure}
     \begin{subfigure}[b]{0.45\textwidth}
         \centering
         \includegraphics[width=\textwidth]{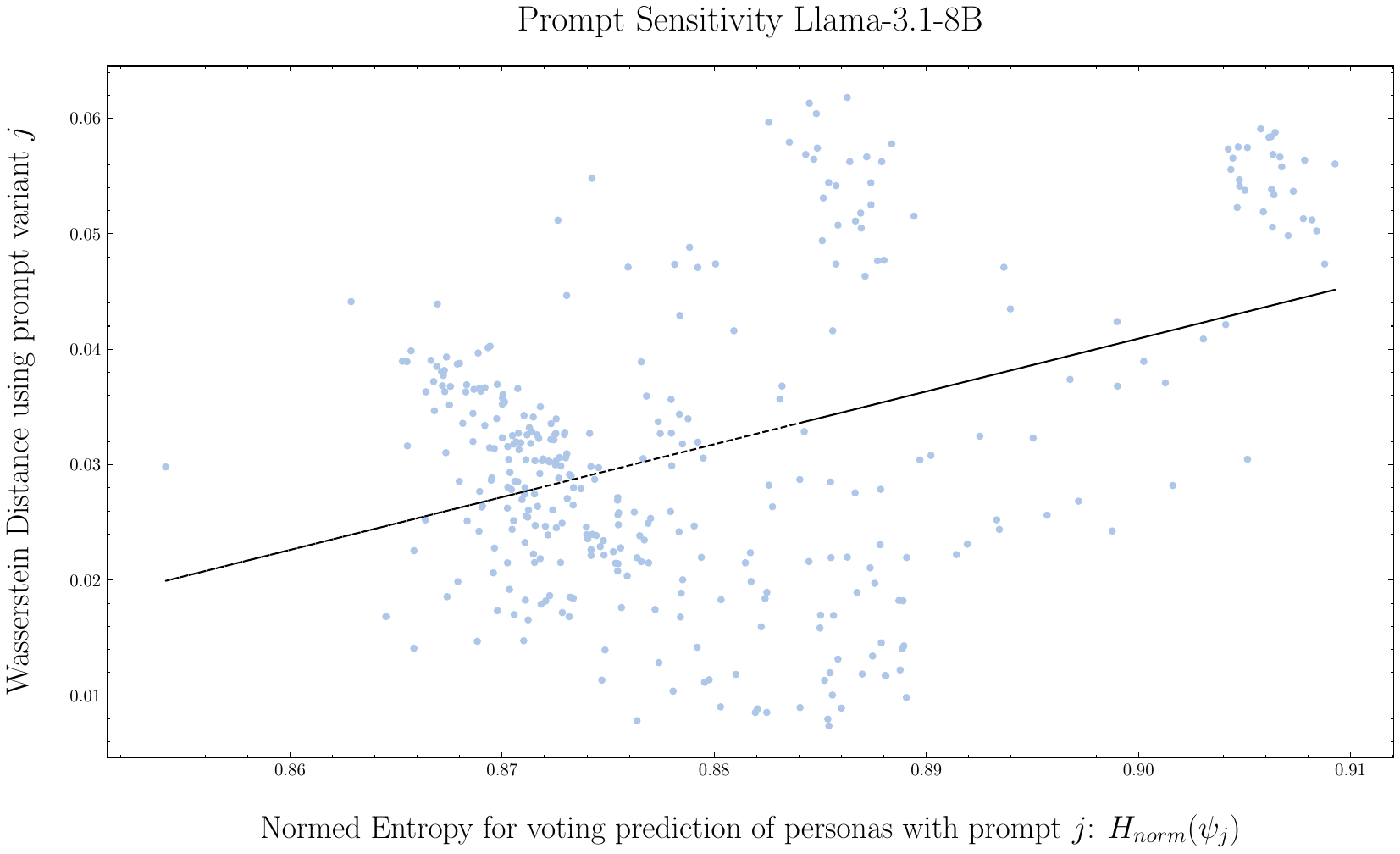}
         \caption{Positive relationship for Llama-3.1-8B-Instruct: higher entropy increases prompt sensitivity.}
         \label{fig:prompt-sensitivity-llama}
     \end{subfigure}
\end{figure}

To analyze prompt sensitivity, we regress the entropy of persona-to-party mappings on the Wasserstein Distance as a proxy for prompt instability. 
The rationale behind this approach is that if minor variations in prompt phrasing significantly alter voting outcome predictions, we should observe a strong relationship between entropy and Wasserstein Distance.
Our results indicate mixed findings across models.
In the case of \textit{Qwen2.5-7B-Instruct} (see~\Cref{fig:prompt-sensitivity-qwen}), we observe a negative relationship: higher entropy in persona-to-party mappings corresponds to lower Wasserstein Distance.
This suggests that when the model exhibits greater uncertainty (higher entropy) in its persona-to-party mappings, it is less sensitive to prompt variations.
In other words, increased entropy appears to stabilize responses across different prompt formulations.
By contrast, the \textit{Llama-3.1-8B-Instruct} model (see~\Cref{fig:prompt-sensitivity-llama}) exhibits a positive relationship.
Here, higher entropy correlates with greater Wasserstein Distance, indicating that when persona-to-party mappings are more uncertain, the model is more affected by prompt variations.
This suggests that for \textit{Llama-3.1-8B-Instruct}, increased entropy amplifies prompt sensitivity, making its voting outcome predictions more unstable under minor prompt reformulations.

%%%%%%%%%%%%%%%%%%%%%%%%%%%%%%%%%%%%%%%%%%%%%%%%%%%%
\section{Discussion and Limitations}

Can LLMs truly replace human surveys for predicting voting outcomes? This study explores how reliably LLMs generate synthetic data by examining persona-to-party mappings in their latent space and prompt sensitivity. 
Our findings question the use of current LLMs for public opinion research, particularly regarding uncertainty in persona associations and variations in model responses due to prompt phrasing.

\textbf{Reliability of LLM-Generated Synthetic Data.}
Our results demonstrate that most LLMs exhibit high entropy in their persona-to-party mappings in their latent space, indicating a broad distribution of voting predictions rather than distinct, well-anchored associations between personas and political preferences.
This is in contrast to real-world voting distributions observed in GLES data, where certain sociopolitical subgroups show more defined voting patterns.
The high entropy in LLM responses suggests that these models inherently introduce a level of uncertainty and dispersion that is not present in actual human survey data.
Interestingly, base models display a pronounced tendency towards right-wing populist preferences, whereas aligned models shift towards center-right and center-left parties.
This shift suggests that alignment processes significantly alter how LLMs interpret and generate survey responses.
The \textit{Qwen2.5-7B} model exhibited the closest match to real-world voting outcomes, yet its latent space entropy did not align perfectly with GLES data, emphasizing that similarity in aggregate predictions does not necessarily imply accurate underlying opinion structures.

Our findings suggest that while LLMs can replicate broad voting trends, they do not accurately capture the demographic-specific distributions of human survey responses (cf. \textit{RQ1}).
This divergence raises concerns about the reliability of synthetic data for opinion research, particularly regarding overgeneralization and potential misinterpretations in model-generated predictions.
% The goal of our first set of experiments is to analyze in how far synthetic data by LLMs can mimic the distribution of human answers in the context of voting outcome prediction. Our results show that the majority of models have a similar and very high distribution entropy across the personas for their persona-to-party mapping in the latent space. This suggests that models display a high degree of uncertainty and that the persona-to-party mapping is more dispersed than what we observe for human respondent groups. 
% When aggregating the predictions for the different personas according to their occurrence in the GLES, we observe noticable differences between base and aligned models. Base models have a predominantly right orientation, while aligning them pushes the models to the CDU (center-right) and more center-left parties. The model, which most closely resembles the outcome distribution of the GLES is ... However, this model is not the one with the most similar entropy in the latent space [??? to be seen]. 

\textbf{Prompt Sensitivity and Stability of Predictions.}
Concerning prompt sensitivity, our analysis reveals key inconsistencies in how LLMs handle slight variations in input phrasing.
While the \textit{Qwen2.5-7B-Instruct} model exhibits a negative relationship between entropy and prompt sensitivity---suggesting that higher uncertainty stabilizes responses---the \textit{Llama-3.1-8B-Instruct} model shows the opposite trend, with greater entropy leading to more instability.
However, beyond these differences, we do not observe a clear or systematic relationship between prompt variations and entropy levels across models. 
This highlights the complexity of LLM behavior and the need for model-specific evaluations when assessing robustness in synthetic survey applications.
These contrasting patterns highlight fundamental differences in how models handle prompt perturbations.

\textbf{Implications for Public Opinion Research.}
LLMs are increasingly used in public opinion research to simulate human preferences~\cite{argyle2023out,bisbee2023synthetic,vonderheyde2024uniteddiversitycontextualbiases}, however, their application presents both opportunities and challenges.
While they can automate surveys, their dispersed persona-party mappings lack structured opinion anchoring, making it difficult to derive reliable insights, particularly for demographic subgroups.
Moreover, high prompt sensitivity means that minor variations in wording can significantly alter results, complicating standardization across studies.
Some models exhibit greater robustness, but others remain highly unstable, limiting their reliability for predictive research.
We caution against uncritical reliance on LLMs as substitutes for human respondents, as their persona-party mappings are often highly dispersed, indicating weakly anchored associations.
This lack of structured alignment reduces confidence in their predictive power.
Future research should focus on refining alignment techniques and probing methodologies to enhance the stability and representational accuracy of synthetic survey responses.

\textbf{Limitations of the Study.} 
While our study provides a comprehensive technical analysis, it is not without limitations.
First, our approach relies on the identification of multi-layer perceptron (MLP) value vectors using trained probes.
Although this method offers valuable insights into how LLMs encode political preferences, it is inherently limited by the accuracy and scope of the probe training process.
Alternative interpretability techniques may yield additional perspectives on model behavior.
Second, our analysis focuses on a selection of LLMs with white-box access. 
The findings may not fully generalize to closed-source models, which might employ different training and alignment strategies.
Future research should examine a broader range of models, including those with different architectures and training data distributions.
Third, our study is constrained by the election context in Germany and the comparison to the year 2021.
While this provides a useful testbed for evaluating LLM reliability in a multi-party setting, different political environments may exhibit distinct dynamics.
Expanding this analysis to other electoral contexts and comparing to more recent election outcomes that are closer to the training data cut-off would enhance the generalizability of our conclusions.

\section{Conclusion}
Our study underscores the challenges and potential pitfalls of using LLMs for opinion research.
While these models can approximate broad trends, their latent space representations and response behaviors diverge significantly from human survey responses.
High entropy in persona mappings, alignment-induced shifts in voting predictions, and prompt sensitivity issues all highlight the need for careful evaluation before deploying LLMs as survey substitutes.
By addressing these limitations through targeted methodological advancements, future research can work towards making AI-generated synthetic data a more reliable tool for public opinion analysis.
\newpage

%Bibliography
\bibliography{main.bib}

\newpage

\appendix
%%%%%%%%%%%%%%%%%%%%%%%%%%%%%%%%%%%%%%%%%%%%%%%%%%%%
\section{Personas}

Overview of persona variables (in german) and corresponding groups used in our study. The table includes demographic variables, political affiliations, and economic factors that define the synthetic personas used for evaluating LLM-generated survey data.

\begin{longtable}{ll}
    \toprule
    \textbf{Parameter} & \textbf{Values} \\ 
    \midrule
    \endfirsthead

    \multicolumn{2}{c}{\textit{(Continued from previous page)}} \\ 
    \toprule
    \textbf{Parameter} & \textbf{Values} \\ 
    \midrule
    \endhead

    \bottomrule
    \endfoot

    \textbf{models} & \begin{tabular}[c]{@{}l@{}}
        meta-llama/Llama-3.1-8B-Instruct, meta-llama/Llama-3.1-8B, \\
        meta-llama/Meta-Llama-3-8B-Instruct, meta-llama/Meta-Llama-3-8B, \\
        meta-llama/Llama-3.2-3B-Instruct, meta-llama/Llama-3.2-3B, \\
        meta-llama/Llama-2-7b-hf, meta-llama/Llama-2-7b-chat-hf, \\
        mistralai/Mistral-7B-Instruct-v0.1, google/gemma-7b-it, \\
        google/gemma-7b, Qwen/Qwen2.5-7B, Qwen/Qwen2.5-7B-Instruct
    \end{tabular} \\ 
    \midrule

    \textbf{parties} & \begin{tabular}[c]{@{}l@{}}
        Alternative für Deutschland, \\
        Christlich Demokratische Union, \\
        Die Linke, \\
        Freie Demokratische Partei, \\
        Bündnis 90/Die Grünen, \\
        Sozialdemokratische Partei Deutschlands
    \end{tabular} \\ 
    \midrule

    \textbf{age} & \begin{tabular}[c]{@{}l@{}}
        jünger als 20, zwischen 20 und 30, zwischen 30 und 40, \\
        zwischen 40 und 50, zwischen 50 und 60, zwischen 60 und 70, \\
        älter als 70
    \end{tabular} \\ 
    \midrule

    \textbf{gender} & weiblich, männlich \\ 
    \midrule

    \textbf{education} & \begin{tabular}[c]{@{}l@{}}
        keinen Abschluss, einen Hauptschulabschluss, \\
        einen Realschulabschluss, Abitur, einen Hochschulabschluss
    \end{tabular} \\ 
    \midrule

    \textbf{hhincome} & niedrig, mittel, hoch \\ 
    \midrule

    \textbf{employment} & \begin{tabular}[c]{@{}l@{}}
        nicht beschäftigt, in Ausbildung, beschäftigt
    \end{tabular} \\ 
    \midrule

    \textbf{left\_leaning} & \begin{tabular}[c]{@{}l@{}}
        stark links, links der Mitte, in der Mitte, \\
        rechts der Mitte, stark rechts
    \end{tabular} \\ 
    \midrule

    \textbf{east\_germany} & Westdeutschland, Ostdeutschland \\ 
    \midrule

    \textbf{year\_of\_election} & 2021, morgen \\
    \bottomrule

\label{tab:variables-and-groups}
\end{longtable}

%%%%%%%%%%%%%%%%%%%%%%%%%%%%%%%%%%%%%%%%%%%%%%%%%%%%
\newpage
\section{Relationship Between Persona Groups and Scaling Factors}
\label{ap:regressionmapping}

\begin{figure}[htbp]
    \centering
    \captionsetup{justification=centering}
    
    \begin{subfigure}{0.48\textwidth}
        \centering
        \includegraphics[width=\linewidth]{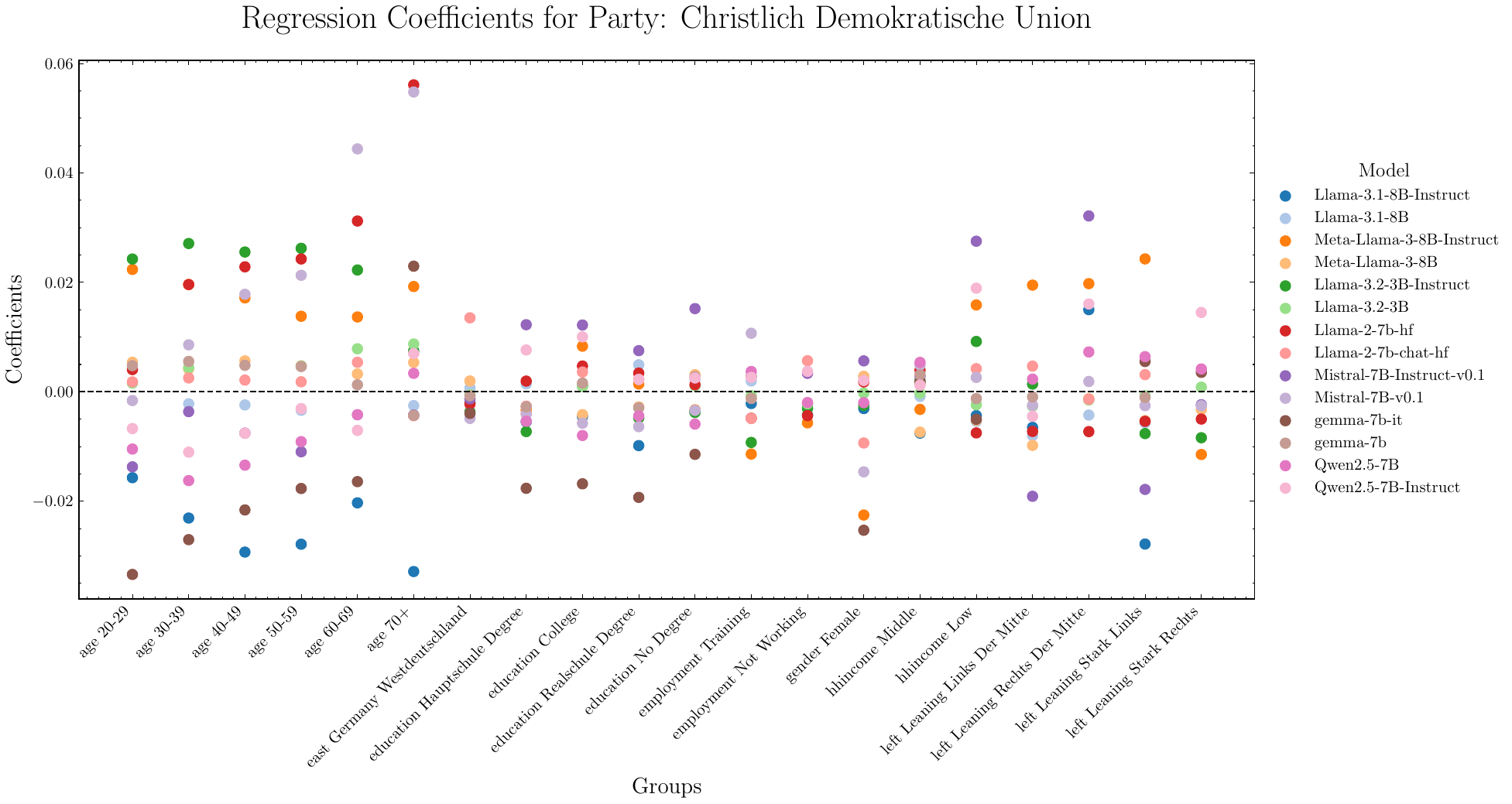}
        \caption{Regression coefficients for CDU}
        \label{fig:persona_cdu}
    \end{subfigure}
    \hfill
    \begin{subfigure}{0.48\textwidth}
        \centering
        \includegraphics[width=\linewidth]{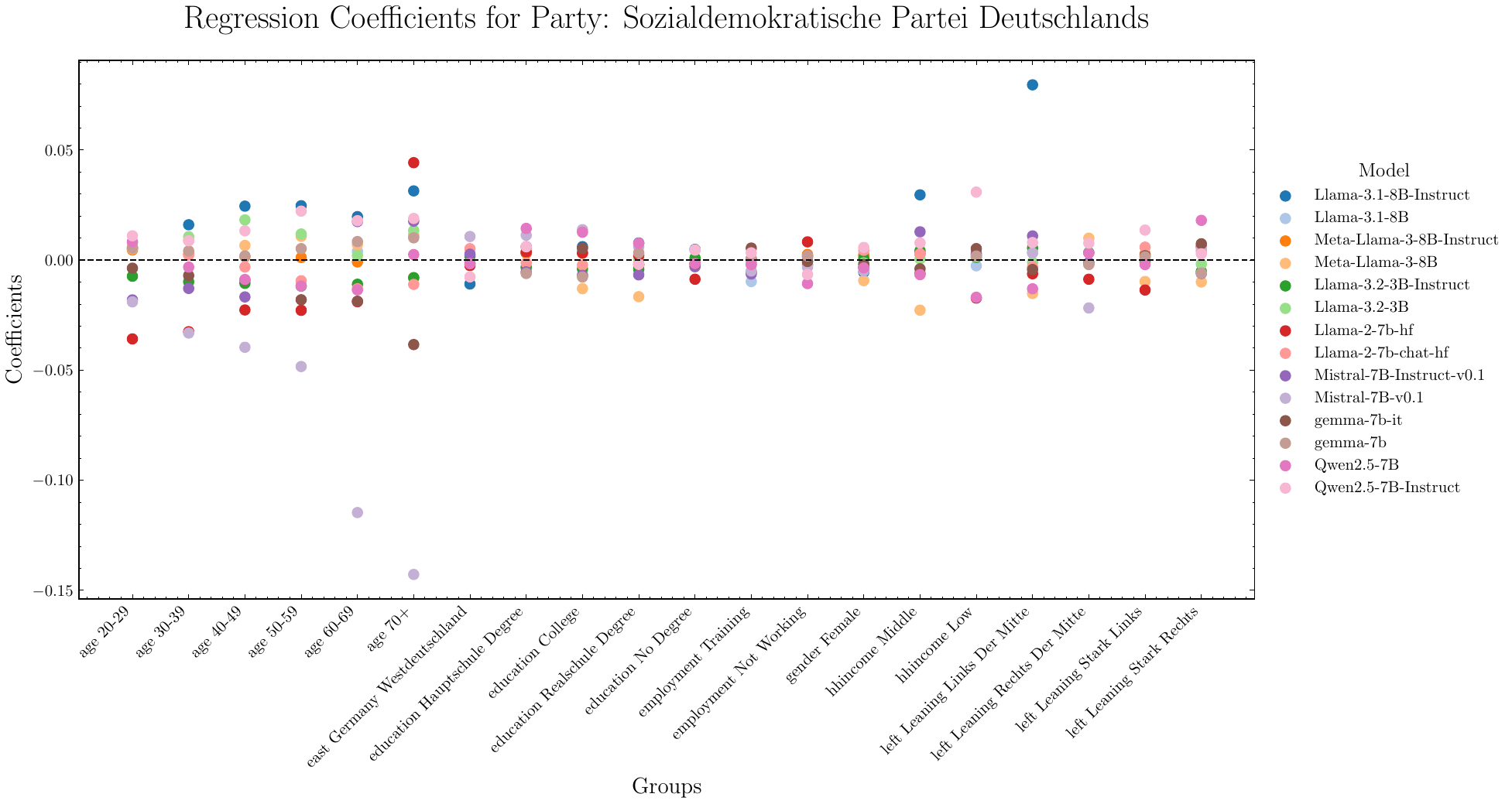}
        \caption{Regression coefficients for SPD}
        \label{fig:persona_spd}
    \end{subfigure}
    
    \vspace{0.4cm}
    
    \begin{subfigure}{0.48\textwidth}
        \centering
        \includegraphics[width=\linewidth]{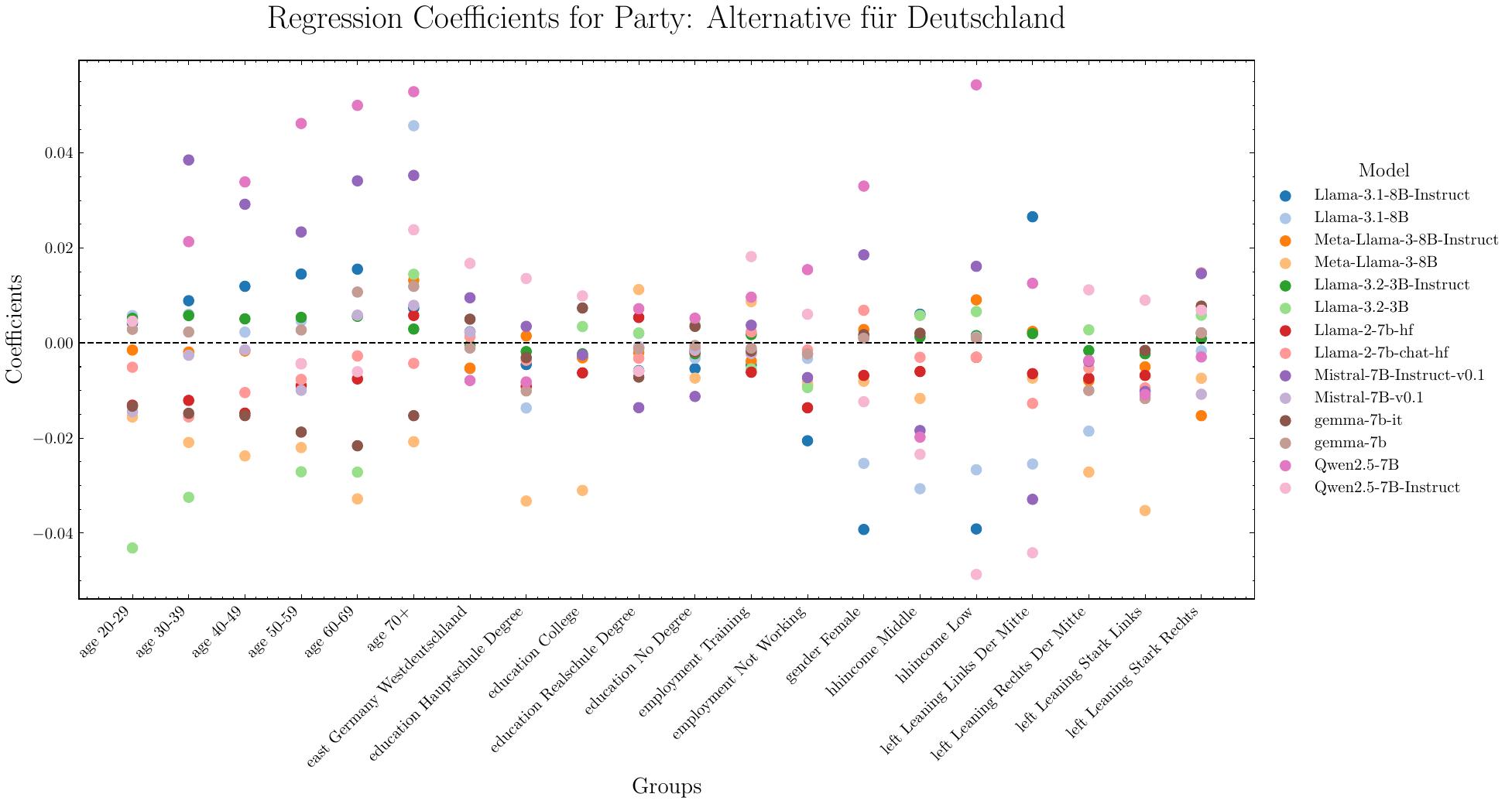}
        \caption{Regression coefficients for AfD}
        \label{fig:persona_afd}
    \end{subfigure}
    \hfill
    \begin{subfigure}{0.48\textwidth}
        \centering
        \includegraphics[width=\linewidth]{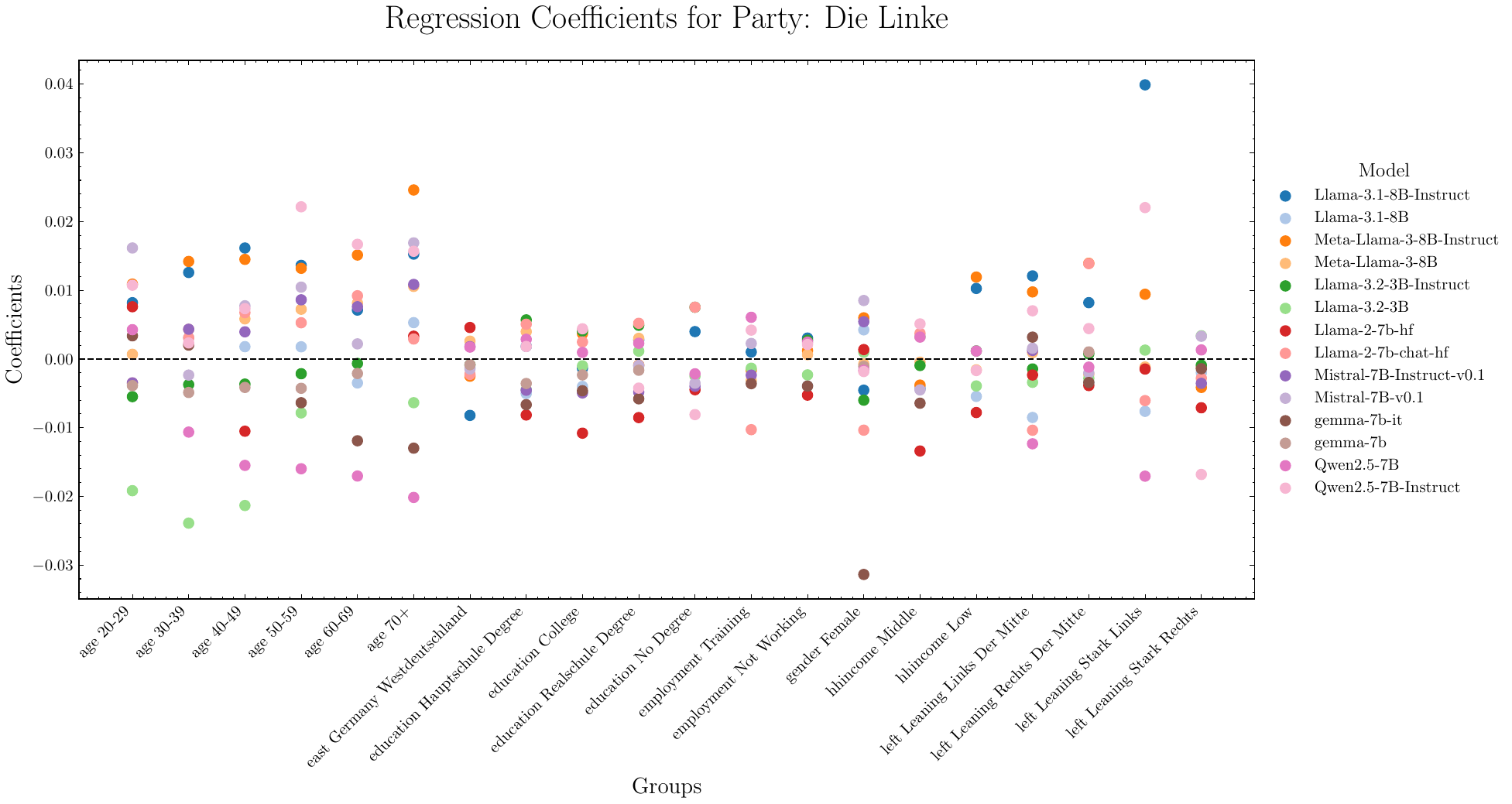}
        \caption{Regression coefficients for Die Linke}
        \label{fig:persona_linke}
    \end{subfigure}
    
    \vspace{0.4cm}

    \begin{subfigure}{0.48\textwidth}
        \centering
        \includegraphics[width=\linewidth]{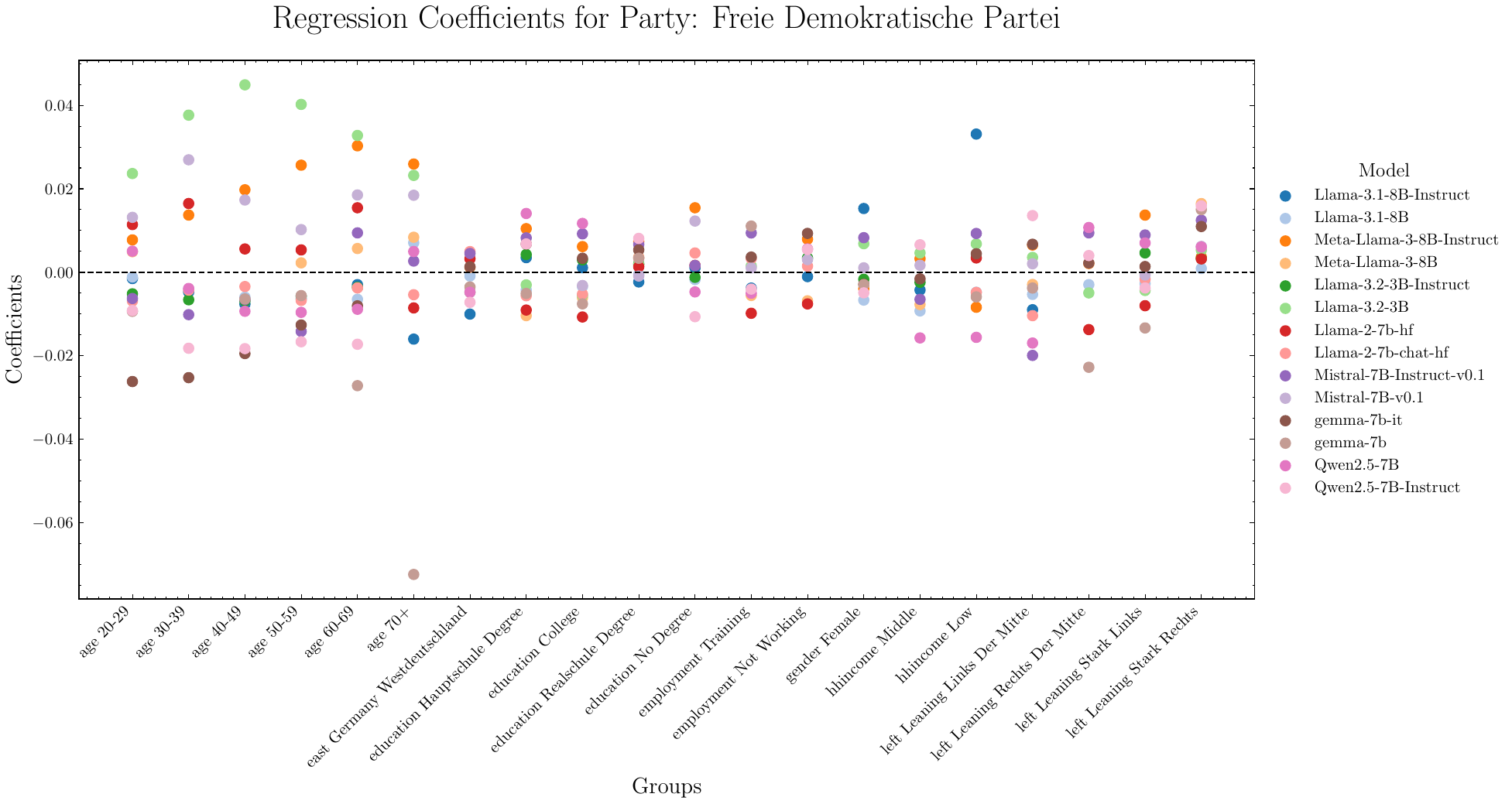}
        \caption{Regression coefficients for FDP}
        \label{fig:persona_fdp}
    \end{subfigure}
    \hfill
    \begin{subfigure}{0.48\textwidth}
        \centering
        \includegraphics[width=\linewidth]{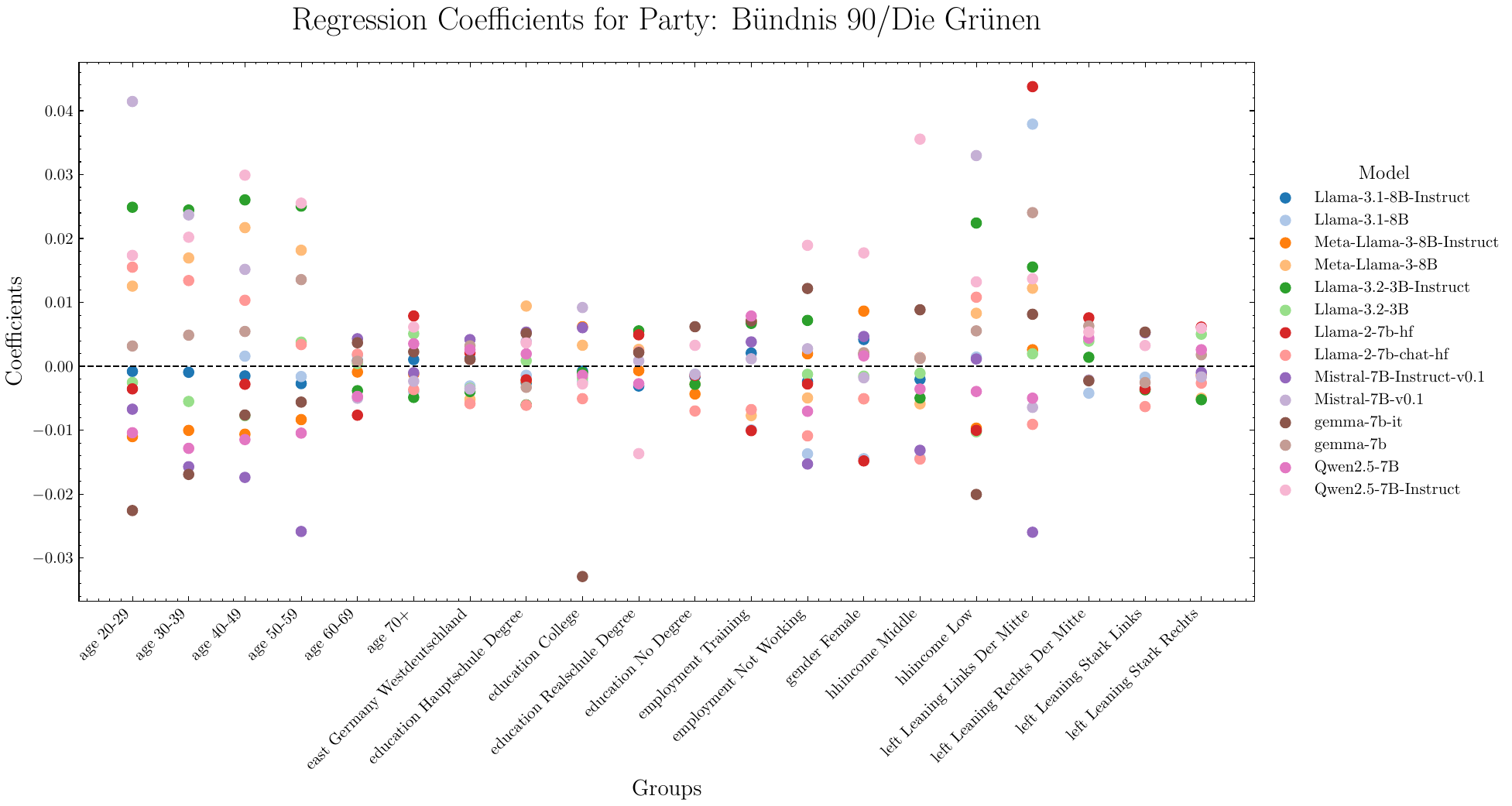}
        \caption{Regression coefficients for Bündnis 90/Die Grünen}
        \label{fig:persona_gruene}
    \end{subfigure}

    \caption{Significant regression coefficients (\( \alpha \leq 0.05 \)) for scaling factors \( m_p^n \) across persona groups \( G \) for each German political party $n$.}
    \label{fig:persona_regression}
\end{figure}

This section presents the significant regression coefficients for scaling factors $m_p^n$ regressed on categorical persona groups $G$, considering a significance level of $\alpha \leq 0.05$. The results are displayed separately for each German political party (see~\Cref{fig:persona_regression}). These coefficients indicate how different persona attributes influence the model's latent space in the form of value vectors across political preferences.

%%%%%%%%%%%%%%%%%%%%%%%%%%%%%%%%%%%%%%%%%%%%%%%%%%%%
\newpage
\section{Entropies}
\label{ap:entropies}

\begin{figure}[h!]
    \centering
    \includegraphics[width=\linewidth]{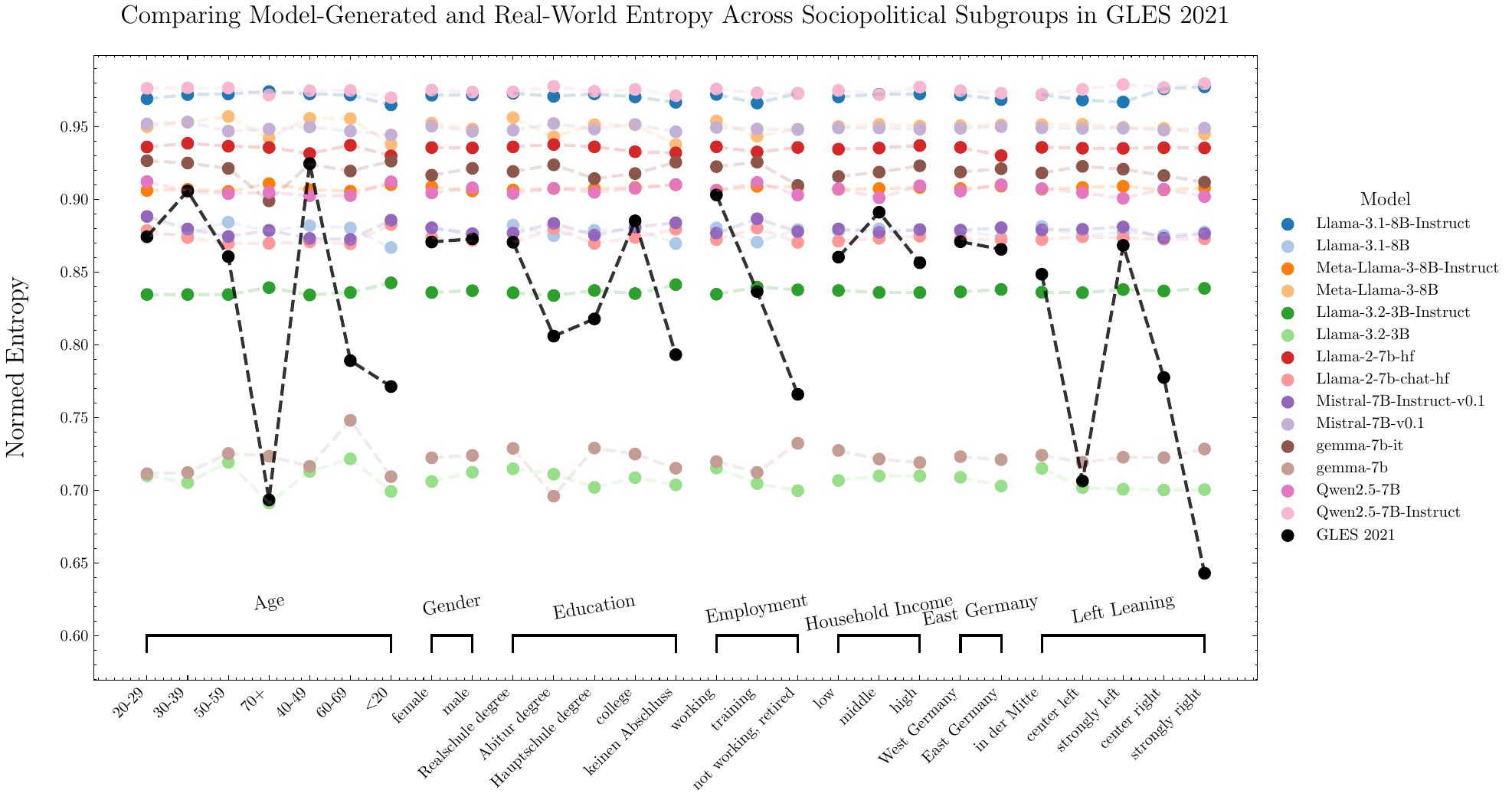}
    \caption{This figure compares the entropy of voting outcomes across different sociopolitical subgroups (e.g., female) as predicted by LLMs versus the real-world entropy observed in the GLES data 2021. The synthetic personas were asked which party they would vote for \textit{tomorrow}, rather than reflecting past election results. Higher entropy indicates greater uncertainty or diversity in political preferences within a subgroup.}
    \label{fig:entropy_tomorrow}
\end{figure}

\begin{figure}
    \centering
    \includegraphics[width=\linewidth]{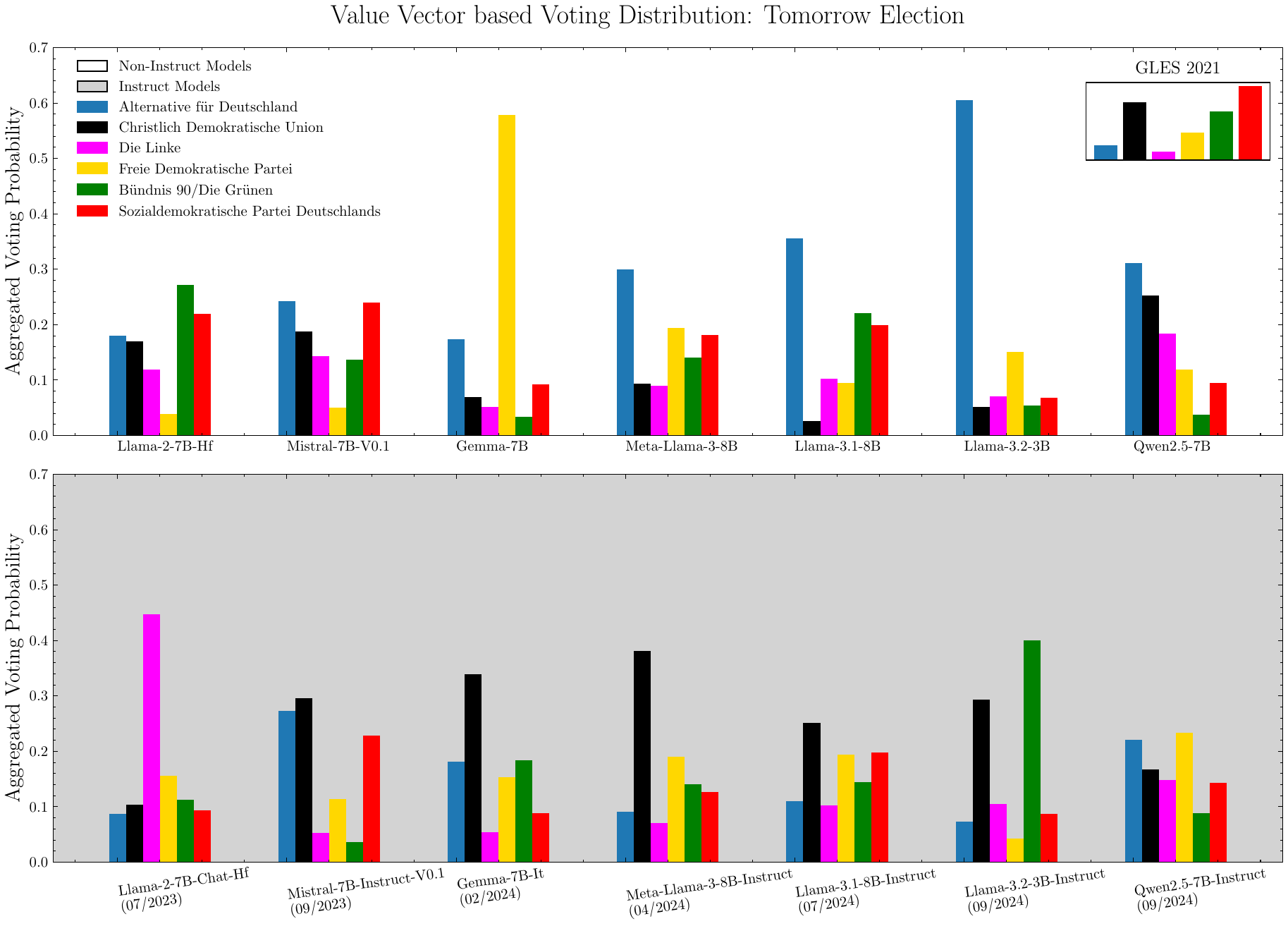}
    \caption{Value vector based distribution $\psi$ in the latent space for election time ``tomorrow'' aggregated over the different personas according to their occurrence in the representative survey GLES.
    The top panel depicts base models, which show a tendency towards the right-populist AfD. The lower panel shows aligned models, where voting distributions shift towards CDU and other democratic, left-leaning parties. \textit{Qwen2.5-7B} is closest to real-world outcomes.}
    \label{fig:voting-dist-tomorrow}
\end{figure}

\end{document}